# ERTIM@MC2: Diversified Argumentative Tweets Retrieval

Kévin Deturck[1], Parantapa Goswami[2], Damien Nouvel[3] and Frédérique Segond[4]

[1] INaLCO ERTIM, 75007 Paris, France/Viseo Innovation, 38000 Grenoble, France
[2] Viseo Innovation, 38000 Grenoble, France
[3] [4] INaLCO ERTIM, 75007 Paris, France

```
kevin.deturck@viseo.com
parantapa.goswami@viseo.com
damien.nouvel@inalco.fr
frederique.segond@inalco.fr
```

**Abstract.** In this paper, we present our participation to CLEF MC2 2018 edition for the task 2 Mining opinion argumentation. It consists in detecting the most argumentative and diverse Tweets about some festivals in English and French from a massive multilingual collection. We measure argumentativity of a Tweet computing the amount of argumentation compounds it contains. We consider argumentation compounds as a combination between opinion expression and its support with facts and a particular structuration. Regarding diversity, we consider the amount of festival aspects covered by Tweets. An initial step filters the original dataset to fit the language and topic requirements of the task. Then, we compute and integrate linguistic descriptors to detect claims and their respective justifications in Tweets. The final step extracts the most diverse arguments by clustering Tweets according to their textual content and selecting the most argumentative ones from each cluster. We conclude the paper describing the different ways we combined the descriptors among the different runs we submitted and discussing their results.

## 1   Introduction

CLEF MC2 Lab 2018 [1] proposes an information retrieval task for festival organizers who would like to know what people think about their event on Twitter[1]. A user's query can be either in French or in English and also specifies a topic from a list of festival names. We design a system, based on linguistic information, which selects the 100 most argumentative and diverse Tweets associated to a user's query. An initial step filters Tweets according to languages and topics in order to reduce the amount of data to be processed. We first extract French and English Tweets performing language detection thanks to an external tool. Then, using regular expressions and key words, a topic filtering step extracts, for each language, sets of Tweets related to the different Festivals.

---
[1] http://www.twitter.com

We perform a linguistic enrichment on the previously extracted sets of Tweets. We then use these Tweets enriched with linguistic information to compute the argumentativity score of each Tweet and measure diversity among Tweets.

Argumentation is a process of construction with arguments that are sets of premises, in other words facts chosen to support claims [2]. Claims are personal statements made by an individual about a topic. Thus, a claim is the expression of an individual's opinion as a polarity (negative, neutral, positive) considering a topic. We link argumentation and opinion in that the former supports the latter. As we said an argumentation is related to an opinion, we measure the argumentativity of a Tweet according the amount of opinion and argumentation it contains. Opinion mining is driven by subjectivity detection because subjectivity is the property of a personal expression and we said opinion is personal. We think the characterization of argumentation by factuality is a crucial marker [3]. Factuality measures how much facts are present in a discourse. A fact is the opposite of subjective content as it stands for a proposition which is true independently of its enunciator. As we mentioned argumentation is a process of construction, we also use discourse structuration markers to detect argumentation.

Diversity is measured on a set of Tweets according to the variety of festival aspects mentioned in the featured view-points. Therefore, the resulting Tweets from our system must be distant considering the aspects they contain. That is why we measure diversity as a distance among Tweets using clustering on their textual content.

In what follows, we present the general architecture of the system together with the different linguistic modules and resources we used. We also explain the different configurations of the runs we have submitted.

## 2    Our approach to the detection of the most argumentative and diverse Tweets within MC2

The overall approach (see Fig.1) consists in applying different filtering steps in order to reduce the original set of "Festival" Tweets to those relevant for the particular task context and to map the most relevant Tweets to user's queries according to their level of argumentativity and diversity.

We reduce the original dataset by two pre-filtering steps to fit the particular task context. The original dataset contains other languages than English and French thus the initial challenge is to identify and to separate English and French Tweets by a language filtering step. A list of festival names is provided as topics for each language. We detect and extract Tweets which contain mentions of these festivals.

We perform data enrichment on the pre-filtered set using Natural Language Processing tools. It consists of morpho-syntactic and semantic information on which the calculation of the argumentativity score is based.

We compute argumentativity score of a Tweet as the amount of both opinion and argumentation it contains. For example, a Tweet with only one claim as "I love Hellfest." will get a lower argumentativity score than a Tweet which combines an opinion and an associated argumentation as in "I love Hellfest because it is ethic.".

We define the diversity of Tweets as the amount of different aspects they mention about the festivals. For example, a set of Tweets about Cannes festival that only contains Tweets like "I love Cannes festival because the introduction was great!" and "Beautiful introduction at Cannes!" is argumentative but not relevant for diversity as it only mentions one aspect. The more diverse Tweets are, the more individuals' critical criteria are provided so that festival organizers get a larger perspective on what people think and why.

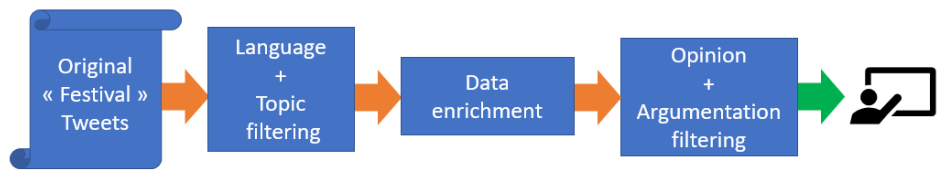

**Fig. 1.** System general architecture

### 2.1 Language filtering

The language filtering step (see Fig. 2) is performed using the Python module "langid.py"[2] ; we choose this module because it combines state-of-the-art results and speed which is essential for processing such a massive dataset [4].

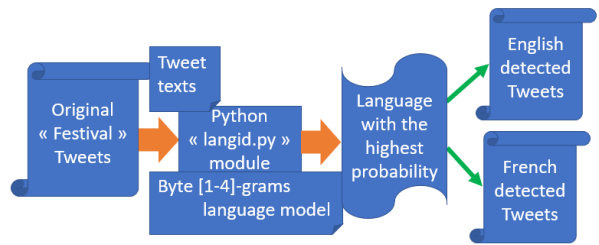

**Fig. 2.** Principle of the language filtering module

---

[2] https://github.com/saffsd/langid.py

## 2.2 Topic filtering

The original dataset contains Tweets that are not only about the festivals from the particular task context. The next step consists, for each language, in detecting and grouping the Tweets into categories corresponding to the lists of festivals provided (see Fig. 3).

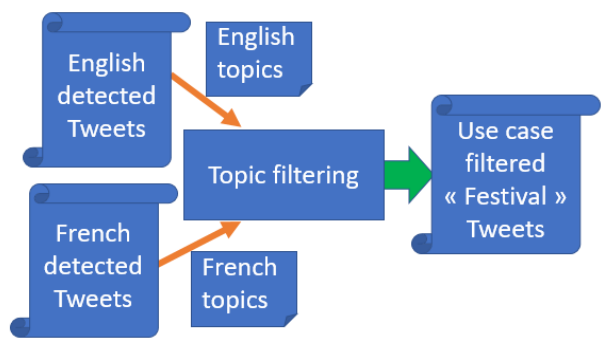

**Fig. 3.** Principle of topic filtering module

Topic detection is performed using regular expressions based on key words representative of each festival. We select a set of "representative" key words associated with each festival based on the mentions in the topically categorized sample of Tweets provided by the organizers. For two festivals, Cannes and Avignon, we notice that the city name is often used alone (without "festival" like "Cannes" instead of "Festival de Cannes") so we decide to only look for "cannes" and "avignon". Regular expressions are built so that the tokens may appear in any case and any order.

## 2.3 Data enrichment

The goal of this intermediary step is to enrich the pre-filtered data with linguistic information. The output of this step is then stored in order to run the process just once preventing lost in performance.

We first normalize each Tweet using the Python module "tweet-preprocessor"[3]. It is fully customizable, allowing us to specify parts of the Tweets we want to remove: URL, Mention, Emoji, Smiley. We decide to keep hashtags as they might contain important information. For example in "The sound is too loud! #FestivalCannes", it allows to identify the topic of the Tweet. We normalize hashtags removing the "#" character.

After text normalization, for each Tweet, we extract the following information using NLP tools we selected according to our needs; they are mostly bilingual and fast to handle the data size (see Table 3).

– List of tokens

---
[3] https://pypi.org/project/tweet-preprocessor/

- List of lemmas
- List of POS labels
- Subjectivity score
- Opinion polarity score

Lists of tokens, lemmas and POS labels are obtained using the TreeTagger tool[4] on the normalized Tweets. We use normalized Tweets because TreeTagger is meant to analyze regular texts while the original Tweets are noisy texts as they can contain Tweet-specific elements like smileys. Specifying as a parameter the language of the text to analyze (English or French), TreeTagger returns a list of lists containing each form, its POS label and lemma.

Subjectivity and opinion polarity scores are obtained using "TextBlob" library[5] and an adaptation for French named "textblob-fr"[6]. TextBlob computes the scores using lexical resources and pattern matching. We run it on the normalized Tweets.

### 2.4 Opinion and argumentation filtering

This step computes an argumentativity score for each Tweet according to the opinion and argumentation it contains. We have selected linguistic features that may represent both aspects.

For opinion detection, we use the subjectivity score as we consider the expression of an opinion to be a subjective content [5]. We consider that the higher its subjectivity score, the more "opinioned" a Tweet is. We also use the opinion polarity score, not for the polarity itself, but for its magnitude that may also indicate how much opinioned a Tweet is. These two scores are combined with their respective weights (specified in section 2.6) as a magnitude score described in equation (2).

$$magnitude(tweet) = \alpha * subjectivity + \beta * |polarity| \qquad (1)$$

where $magnitude(tweet)$ is the opinion magnitude score (comprised between [0-1]) for $tweet$, $subjectivity$ the subjectivity score (comprised between [0-1]), $polarity$ the polarity score in [0-1], $\alpha$ and $\beta$ are their respective weights

We also use two lexical resources: one for English and one for French. For English, we use [6] which encodes the "arousal" property of 13,915 English lemmas. It associates a score to each lemma according to the affectivity it denotes; our hypothesis is that the more a Tweet contains high affectivity lemmas (high scores), the more opinioned it would be (see equation 3). For French, we use [7], a French lexicon which associates to 14,129 non-neutral lemmas a binary polarity value ("positive" or "negative") and six binary values depending on whether each lemma evokes (1) or not (0) a sentiment among six different: joy, anger, surprise, sadness, disgust and fear. We consider sentiment as an internal psychologic state whose expression can serve the formulation of an opinion. Our hypothesis is that the more a French Tweet contains lemmas present in this lexicon (encoding only non-neutral lemmas) and with high number of sentiment denotations, the more opiniated it would be (see equation 4).

---

[4] http://www.cis.uni-muenchen.de/~schmid/tools/TreeTagger/
[5] http://textblob.readthedocs.io/en/dev/
[6] https://github.com/sloria/textblob-fr

$$arousal(tweet) = \frac{\sum_{i=1}^{n} arousal(lemma_i)}{n} \quad (2)$$

where $arousal(tweet)$ is the arousal score comprised between [0-1] for $tweet$, $arousal(lemma_i)$ the lexicon-based arousal score normalized comprised between [0-1] for the lemma $lemma_i$ and $n$ the number of lemmas in $tweet$

$$expressivity(tweet) = \frac{\sum_{i=1}^{n} expressivity(lemma_i)}{n} \quad (3)$$

where $expressivity(tweet)$ is the expressivity score comprised between [0-1] for $tweet$, $expressivity(lemma_i)$ is the expressivity of $lemma_i$ computed following equation (5) and $n$ is the number of lemmas in $tweet$

$$expressivity(lemma) = \frac{|true|}{7} \quad (4)$$

where $expressivity(lemma)$ is the lexicon-based expressivity score for $lemma$ according to $|true|$, the number of valid properties among the presence in the lexicon and the six lexicon-annotated sentiments

Besides these lexicon-based measures, we also detect the opinion in a Tweet by taking into account the proportion of adjectives regarding POS tags; our hypothesis is that the more a Tweet contains adjectives, the more opiniated it would be (see equation 6).

$$descriptivity(tweet) = \frac{|adjectives|}{n} \quad (5)$$

where $descriptivity(tweet)$ is the descriptivity score for $tweet$ according to $|adjectives|$, the number of tokens tagged as adjectives and $n$ the number of tokens in $tweet$

Regarding argumentation, we say that an argumentative text is particularly structured to effectively combine arguments and opinions. Conjunctions are discourse connectors thus we suppose they are particularly used to structure a text. We use POS tags to value the proportion of conjunctions in a Tweet (see equation 7).

$$structuration(tweet) = \frac{|conjunctions|}{n} \quad (6)$$

where $structuration(tweet)$ is the structuration score comprised between [0-1] for $tweet$, $|conjunctions|$ the number of conjunctions in $tweet$ and $n$ the number of tokens in $tweet$

For English Tweets, we compute a concreteness score (see equation 8) relying on the lexical resource [8]. It associates to nearly 40,000 English lemmas a score which indicates how much perceptible (by the five senses) their mean is. As we start from the hypothesis that an argumentative text is factual, thus independent from an individual's state of mind, we formulate a new hypothesis saying that it may contain more concrete lemmas.

$$concreteness(tweet) = \frac{\sum_{i=1}^{n} concreteness(lemma_i)}{n} \quad (7)$$

where $concreteness(tweet)$ is the concreteness score comprised between [0-1] of $tweet$, $concreteness(lemma_i)$ the lexicon-based concreteness score of $lemma_i$ (normalized between [0-1], 0 if lemma is missing) and $n$ the number of tokens in $tweet$

### 2.5 Diversity filtering

In this final step, we build a set of Tweets that maximizes the diversity criterion among the most argumentative Tweets. Diversity measures how much festival aspects the Tweets mention. Thus, we suppose diverse Tweets may contain words semantically distant. For example, we detect that "This festival is too expensive." and "Ticket price for this festival is too high." mention a similar aspect by the semantic proximity between the words "expensive" and "price". Inversely, the texts "This festival program is so good!" and "This festival proposes a good choice of beers!" are more distant due to the semantic distance between the words "program" and "beer".

As diversity is computed according to the lexical semantics distance between Tweets, we use word embeddings models from Sketch Engine[7] to get a spatial representation of words, one for English and one for French. As we want to keep as much form-wise information as possible, we select for both languages word form models (without lowercasing). For English, we select the model based on British National Corpus because it is the lighter therefore it avoids memory problem at loading time. For French, the only one proposed is a model based on a Web corpus. We vectorize a Tweet by matching its tokens against the model using FastText Python module[8]. We use K-means clustering via the ScikitLearn toolkit[9] to compute the distance between vectorized Tweets.

### 2.6 Impacts of dataset pre-filtering

Table 1 shows the efficiency of the pre-filtering steps (regarding languages and topics) in reducing data size by evaluating compression ratios among selective properties for argumentativity. It includes linguistic properties (lemmas, subjectivity and opinion polarity) obtained as described in section 2.3. We consider properties which are relative indicators of point of view diversity in our data regarding sources (authors) and vocabulary (lemmas). We also select subjectivity and opinion polarity scores as a mean to measure inclination of authors to express and explain their point of view.

---

[7] https://embeddings.sketchengine.co.uk/static/index.html  
[8] https://github.com/facebookresearch/fastText/tree/master/python  
[9] http://scikit-learn.org/stable/index.html

The language filtering step removes more than 40% of the original data compressing the unique authors ratio, computed with equation (8), by more than 50% for both languages. The dataset is much more practicable but with an impoverishment of sources. Only around 1% of the lemmas used in Tweets are different, which may be difficult for lexical approaches. Polarity and subjectivity average magnitudes ([0-1] interval scale) are low among the two languages; it may be positive to distinguish argumentative Tweets.

$$ratio = \frac{nUniqAuthors}{nTweets} \tag{8}$$

where $ratio$ is the ratio of the number of unique authors, $nUniqAuthors$ to number of Tweets $nTweets$

We can observe in Table 1 the evolution of author and vocabulary usage between the two first filtering steps. Unique authors ratio increases by around 80% for both languages; it is a considerable increase compared to the previous step and a positive result for the representativeness of the data. Vocabulary usage is poor in English (0.2% of different lemmas) and in French (0.5%); this may be a relevant concern for the diversity criterion. Polarity and subjectivity average magnitude stay low even if it increases for French Tweets; the selective power of this information may be preserved.

**Table 1.** Statistics on dataset through the pre-filtering steps

| Language | Initial multilingualism | English | | French | |
|---|---|---|---|---|---|
| Dataset | Initial "Festival" Tweets | Language filtered | Topic filtered | Language filtered | Topic filtered |
| #Tweets | 63M | 34M | 2M | 3M | 200k |
| #Unique authors | 45M | 9M | 1M | 1M | 100k |
| #Tokens[10] | 960M | 532M | 25M | 41M | 3M |
| #Unique lemmatized tokens[11] | N/A[12] | 7M | 61k | 252k | 15k |
| Subjectivity magnitude average[13] | N/A[14] | 0.28 | 0.28 | 0.26 | 0.15 |

---

[10] Obtained using Unix 'wc' command
[11] Lemmatized using http://www.cis.uni-muenchen.de/~schmid/tools/TreeTagger/
[12] [14] Not designed to support all languages of the original dataset
[13] Obtained using https://github.com/sloria/textblob

| | | | | | |
|---|---|---|---|---|---|
| Polarity magnitude average[15] | N/A[16] | 0.18 | 0.14 | 0.13 | 0.07 |

## 2.7 Runs description

A run returns the 100 most argumentative and diverse Tweets for all languages and festival names from the particular task context. To get the 100 most argumentative and diverse Tweets, we run K-Means with k = 100 and select the Tweet with higher argumentativity score from each cluster. Each run results in a ranked set of Tweets with the most argumentative first.

We have submitted three runs which differ by features and associated weights used for computing the argumentativity score of each Tweet. We combine scores, described in section 2.4, that use the same types of linguistic information (POS-based and lexical types). The purpose is to evaluate the impact of Mathematically, the combination is an arithmetic mean (see equations 9, 10 and 11).

$$posScore(tweet) = \frac{structuration(tweet) + descriptivity(tweet)}{2} \quad (9)$$

$$lexicalScoreEn(tweet) = \frac{arousal(tweet) + concreteness(tweet)}{2} \quad (10)$$

$$lexicalScoreFr(tweet) = expressivity(tweet) \quad (11)$$

In all runs, we set weights in magnitude equation (2) with $\alpha = 0.75$ and $\beta = 0.25$ as we are more confident in the subjectivity analysis than in the polarity one, relying on our observations of some analyzed data. Even if we do not have a lot of confidence in the magnitude score according to our observations of some analyzed data, it is the feature that is the more directly related to argumentativity. Therefore we use it in all runs mostly with a minor weight (0.25). Run 1 uses all types of features while the two others interchange the uses of lexicon-based and POS-based scores to evaluate their respective impact on result quality.

Run 1 uses all types of feature: POS, opinion score and the lexicon-based score (see equations 12 and 13). English lexical resources cover opinion and argumentation aspects whereas the French one only covers opinion. We decide to give a minor weight in the French run (0.25) than in the English one (0.50). We try to balance the lack of lexicon resource for argumentation in French giving a more important weight to POS-based score (0.50) in comparison with English (0.25). The magnitude score gets the same weight for the two languages (0.25) as we do not have a lot of confidence in the tool that computes the score.

---

[15] Obtained using https://github.com/sloria/textblob
[16] Not designed to support all languages of the original dataset

$$argumentativityEn(tweet) = 0.25\ magnitude(tweet) + 0.50 * lexicalScoreEn(tweet) + 0.25 * posScore(tweet) \quad (12)$$

$$argumentativityFr(tweet) = 0.25 * magnitude(tweet) + 0.25 * lexicalScoreFr(tweet) + 0.50 * posScore(tweet) \quad (13)$$

Run 2 uses the magnitude and lexicon-based scores (see equations 14 and 15). As we previously said, the French lexicon coverage of task aspects is not complete contrary to the English one so we set it with a smaller weight in French (0.50) than in English (0.75). In English, we give a major weight to the lexicon-based score because we attach more importance to the manually built lexical resource in comparison with the magnitude score automatically computed.

$$argumentativityEn(tweet) = 0.25 * magnitude(tweet) + 0.75 * lexicalScoreEn(Tweet) \quad (14)$$

$$argumentativityFr(tweet) = 0.50 * magnitude(tweet) + 0.50 * lexicalScoreFr(tweet) \quad (15)$$

Run 3 uses the POS-based score in association with the magnitude score (see equation 16). As the tool we used to extract the POS labels is the same for English and French, we give the same weight (0.75) to POS-based score for the two languages. It is a major score because of the lack of reliability we have for the opinion score.

$$argumentativity(tweet) = 0.25 * magnitude(tweet) + 0.75 * posScore(tweet) \quad (16)$$

## 3   Results, conclusions and perspectives

An overview of the results obtained by the different systems in the lab can be found in [9]. In this section, we logically focus on our system as it is the subject matter of the present paper and because we are in a relevant position to review its results. At this time, no diversity results have been provided by the organizers.

### 3.1 Results

Regarding argumentativity, the organizers used two measures to evaluate the quantity of argumentative content in runs. NDCG measures the relevance of Tweets with a discount function over the rank: in each run, the most relevant Tweets must appear first. NDCG measures the relevance of each run according to regular expressions which match argumentative content. Two references for argumentative content have been used: one manual prepared by annotators and another one obtained by a pooling of runs from the different participants' systems. A measure named "%arg" gives the percentage of argumentative content comparing to both pooling and manual references. Table 2 presents the results for all runs.

**Table 2.** Argumentativity results

| Language | English | | | French | | |
|---|---|---|---|---|---|---|
| Measure | NDCG-manual | NDCG-pooling | %arg | NDCG-manual | NDCG-pooling | %arg |
| Run1 | 0.002 | 0.36 | 21.81 | **2.597** | **2.06** | **22.00** |
| Run2 | **0.007** | **0.60** | **36.72** | 2.594 | 1.39 | 20.43 |
| Run3 | 0.003 | 0.39 | 20.36 | 2.594 | 1.99 | 21.89 |

We observe in Table 2 the best results are not obtained with the same types of features across the languages. All runs use the magnitude of subjectivity and polarity scores (see section 2.7) but in English the best results are obtained by addition with lexical features (run 2) while in French the best run combines lexical and POS-based features (run 1). We explain this difference by the different natures of lexical resources across the languages; as we suspected preparing the runs (see section 2.7), the lexical resource for English may be more related to argumentativity especially with the concreteness property while the French one is only about sentiment expression which may be useful for opinion mining (see section 2.4) but not sufficient to detect argumentative content.

Comparing the results in one language, run 2 in French is particularly low (NDCG-pooling and %arg). This run in French may not include enough features related to argumentativity; the presence of opinion polarity, subjectivity or sentiments in a Tweet should indicate that it contains a personal expression but it does not imply that it is justified by an argumentation. However, the addition of the lexical sentiment feature allows run 1 to be the best in French (in comparison with run 3). We think that a personal content may be the base for argumentation as a supporting tool. In other words, particularly on Twitter, we suppose that there might not be argumentation without a personal content. We note that POS-based information considering the structuration is effective even on Tweets, probably due to relativity of POS-based scores among Tweets. In English, it is surprising to observe that the run with lexical and POS-based features (run 1) gets lower results than the run without POS-based information (run 2). Regarding the weights among the two runs (see section 2.7), the lexical feature is ¾ of the score in run 2 whereas it is ½ in run 1; we think that the lower results in run 1 compared to run 2 might be explained by the lower importance of the lexical feature in run 1 rather than the addition of the POS-based feature. It is supported by the result of run 3 which uses the POS-based feature and gets a better NDCG score than run 1.

Considering our position across the different participants' systems is interesting because we are at the first place by pooling for both languages (lowest scores are 0.00 in French and 0.05 in English) and at the last place for English (best score is 0.06) and penultimate for French (the lowest score is 2.28 from the baseline and the best score is 2.89). It means that our system does not correctly match the manual reference but extracts arguments not considered by the annotators or other participants. Maybe it reflects a divergence in what is considered as argumentative.

### 3.2 Conclusions and perspectives

The hypothesis of argumentation using words which denote concrete things seems to be validated by the importance of the corresponding lexical feature in English, getting a better score when it is used with a greater weight. In French, the discourse connectors feature gives the best results and validates the assumption of a more structured text when it is argumentative, even on short messages like Tweets.

As the lexicon encoding concreteness and arousal properties allows to get the best results, it may be relevant to build a corresponding resource in French; it could be achieved by a translation process. It would be interesting to view if we also get better results in French.

We would like to analyze similar features on other text media to compare their respective contribution. In particular, it would be interesting to evaluate the POS-based feature with structuration words on texts which are less bound by their size, maybe a longer text needs to be more structured.